\documentclass[letterpaper, 10pt, conference]{ieeeconf}  % Comment this line out if you need a4paper

\IEEEoverridecommandlockouts                              % This command is only needed if 
\usepackage{algorithm}
\usepackage[noend]{algpseudocode}

\usepackage[noadjust]{cite}

\makeatletter
\let\NAT@parse\undefined
\makeatother

\usepackage{amsmath,amsfonts,bm}

\def\eqref#1{equation~\ref{#1}}
\def\1{\bm{1}}

\def\rva{{\mathbf{a}}}

\def\rvo{{\mathbf{o}}}

\def\rvs{{\mathbf{s}}}

\def\rmA{{\mathbf{A}}}

\def\rmI{{\mathbf{I}}}

\DeclareMathAlphabet{\mathsfit}{\encodingdefault}{\sfdefault}{m}{sl}
\SetMathAlphabet{\mathsfit}{bold}{\encodingdefault}{\sfdefault}{bx}{n}

\def\gL{{\mathcal{L}}}

\def\gN{{\mathcal{N}}}

\newcommand{\E}{\mathbb{E}}

\newcommand\norm[1]{\left\lVert#1\right\rVert}

\usepackage[dvipsnames]{xcolor}

\usepackage{kotex}
\usepackage{lipsum}
\usepackage{pifont}% http://ctan.org/pkg/pifont
\usepackage[normalem]{ulem}
\usepackage{mathtools}

\usepackage{array}
\newcolumntype{P}[1]{>{\centering\arraybackslash}p{#1}}
\newcolumntype{M}[1]{>{\centering\arraybackslash}m{#1}}
\newcolumntype{L}[1]{>{\hspace{0.5em}\raggedright\arraybackslash}m{#1}}
\newcolumntype{R}[1]{>{\raggedleft\arraybackslash}m{#1}}

\usepackage{xspace}

\makeatletter
\DeclareRobustCommand\onedot{\futurelet\@let@token\@onedot}
\def\@onedot{\ifx\@let@token.\else.\null\fi\xspace}

\def\eg{\emph{e.g}\onedot} 
\def\ie{\emph{i.e}\onedot}

\makeatother

\usepackage[colorlinks=true, linkcolor=blue, citecolor=blue, urlcolor=blue, pagebackref=true]{hyperref}
\definecolor[named]{ACMDarkBlue}{cmyk}{1,0.58,0,0.21}
\definecolor{darkgreen}{RGB}{25,200,25}
\hypersetup{colorlinks,
  linkcolor=ACMDarkBlue,  % ACMPurple,
  citecolor=ACMDarkBlue,  % ACMPurple,
  urlcolor=ACMDarkBlue,   % black,
  filecolor=ACMDarkBlue
}

\usepackage{cleveref}
\Crefformat{figure}{Fig.~#2#1#3}
\Crefformat{equation}{(#2#1#3)}
\Crefformat{table}{Tab.~#2#1#3}

\usepackage{caption}
\DeclareCaptionFormat{tablewithgap}{#1#2\vspace{3pt}#3}
\usepackage{tabularx,booktabs}
\newcolumntype{Y}{>{\centering\arraybackslash}X}
\usepackage{colortbl}
\usepackage{multirow}
\usepackage{soul}
\usepackage{wrapfig}

\usepackage{footmisc}
\newcommand{\dagfootnote}[1]{%
  \textsuperscript{\dag}% 본문 마커
  \begingroup
    \let\savedfootnoterule\footnoterule
    \renewcommand{\thefootnote}{}% 번호 숨김
    \renewcommand{\footnoterule}{% 이 각주에만 가로선
      \vspace{1ex}\hrule width 0.4\textwidth\vspace{1ex}%
    }%
    \footnotetext{\dag\ #1}% 아래쪽 각주 텍스트
    \addtocounter{footnote}{-1}% 전체 번호에 영향 없게 롤백
    \let\footnoterule\savedfootnoterule
  \endgroup
}

\usepackage{algorithm,algpseudocode}
\usepackage[algo2e]{algorithm2e} 
\definecolor{commentcolor}{RGB}{110,154,155}
\definecolor{defcolor}{RGB}{225,81,145}
\usepackage{titletoc}

\makeatletter
\let\labelindent\relax
\makeatother

\usepackage{enumitem}
\usepackage{flushend}

\definecolor{darkgreen}{RGB}{25,200,25}

\definecolor{revcolor}{RGB}{245,135,0}

\definecolor{acg_blue}{RGB}{0, 90, 190}

\newcommand{\method}{ACG\xspace}

\title{\LARGE \bf
ACG: Action Coherence Guidance for Flow-based Vision-Language-Action models
\vspace{-0.8cm}
}

\author{Minho Park$^*$, Kinam Kim$^*$, Junha Hyung, Hyojin Jang, Hoiyeong Jin, Jooyeol Yun, Hojoon Lee$^\dagger$, Jaegul Choo$^\dagger$\\%
DAVIAN-Robotics, KAIST AI 
\thanks{$^*$Equal contribution, $^\dagger$Corresponding authors.}%
\thanks{
\textbf{Links:}
\href{https://github.com/DAVIAN-Robotics/ACG}{Github Code}
$\mid$ \href{https://huggingface.co/collections/DAVIAN-Robotics/acg-gr00t-n1-2b-post-trained-models}{HF Models}
$\mid$ \href{https://DAVIAN-Robotics.github.io/ACG}{Project page}
}%
}

\begin{document}

\maketitle

\thispagestyle{empty}
\pagestyle{empty}

%%%%%%%%%%%%%%%%%%%%%%%%%%%%%%%%%%%%%%%%%%%%%%%%%%%%%%%%%%%%%%%%%%%%%%%%%%%%%%%%
\begin{abstract}
Diffusion and flow matching models have emerged as powerful robot policies, enabling Vision-Language-Action (VLA) models to generalize across diverse scenes and instructions.
Yet, when trained via imitation learning, their high generative capacity makes them sensitive to noise in human demonstrations: jerks, pauses, and jitter which reduce action coherence. Reduced action coherence causes instability and trajectory drift during deployment, failures that are catastrophic in fine-grained manipulation where precision is crucial.
In this paper, we present Action Coherence  Guidance (ACG) for VLA models, a training-free test-time guidance algorithm that improves action coherence and thereby yields performance gains.
Evaluated on RoboCasa, DexMimicGen, and real-world SO-101 tasks, ACG consistently improves action coherence and boosts success rates across diverse manipulation tasks.
\end{abstract}

% \textbf{Links:}
% \href{https://github.com/DAVIAN-Robotics/ACG}{Github Code}
% $\mid$ \href{https://huggingface.co/collections/DAVIAN-Robotics/gr00t-n1-2b-post-trained-models}{HF Models}
% $\mid$ \href{https://DAVIAN-Robotics.github.io/ACG}{Project page}
%%%%%%%%%%%%%%%%%%%%%%%%%%%%%%%%%%%%%%%%%%%%%%%%%%%%%%%%%%%%%%%%%%%%%%%%%%%%%%%%

\section{Introduction}

Diffusion and flow matching models are reshaping how robots learn to manipulate objects~\cite{chi2023diffusion}.
These generative models act as robot policies that directly model complex action distributions from human demonstrations, enabling strong generalization across diverse manipulation tasks.
This paradigm has been further extended to Vision-Language-Action (VLA) models, enabling generalization across a wide range of scenes and language     instructions~\cite{bjorck2025gr00t, black2024pi_0, shukor2025smolvla, liu2025rdt}.

Despite these advances, diffusion and flow matching policies trained via imitation learning remain highly sensitive to noise in human demonstrations, such as pauses, jerks, or jitter~\cite{mandlekar2021matters, fang2025demonstration, yuan2023good, tangkaratt2021robust}.  
Their large generative capacity often memorizes these imperfections, which degrades \textit{action coherence} of the learned policies ~\cite{act, liu2025bidirectional}. 
Formally, \textit{action coherence} denotes the smoothness and consistency of successive actions, which can be measured by variability or jerks~\cite{rudin1992nonlinear,flash1985coordination}.

During deployment, the loss of action coherence leads to two key failures. First, unstable actions can cause instability at critical moments. For example, in a pick-and-place task, the robot may fumble near the target object or inadvertently push it away. Second, even minor action noise can accumulate over time, causing the robot’s trajectory to drift from desired states. Therefore, enhancing action coherence between action sequences is essential for ensuring reliable and robust manipulation.

\begin{figure}
\centering
\includegraphics[width=\linewidth]{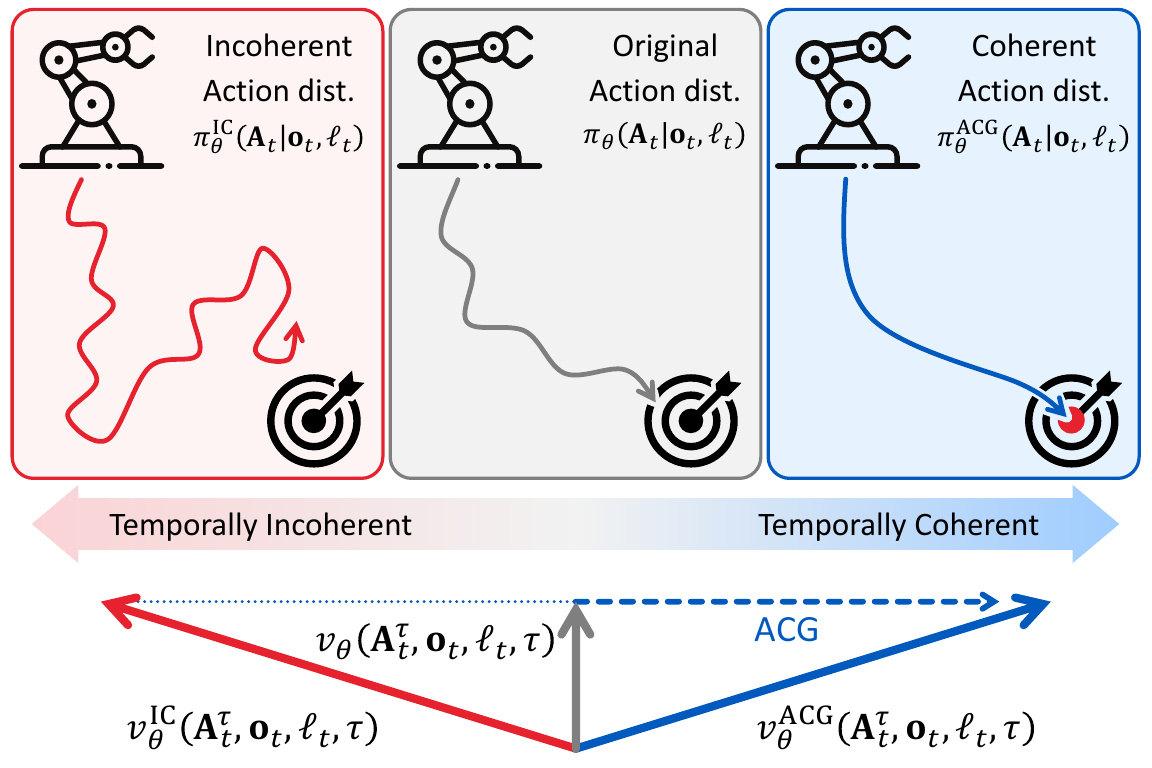}
\caption{Conceptual illustration of \method. \method constructs an incoherent vector field $v_\theta^\text{IC}$ and combines it with the original vector field $v_\theta$ to extrapolate a guidance vector that steers sampling toward coherent action sequences.}
\label{fig:1_conceptual_figure}
\vspace{-0.3cm}
\end{figure}

\begin{figure}
\centering
\includegraphics[width=\linewidth]{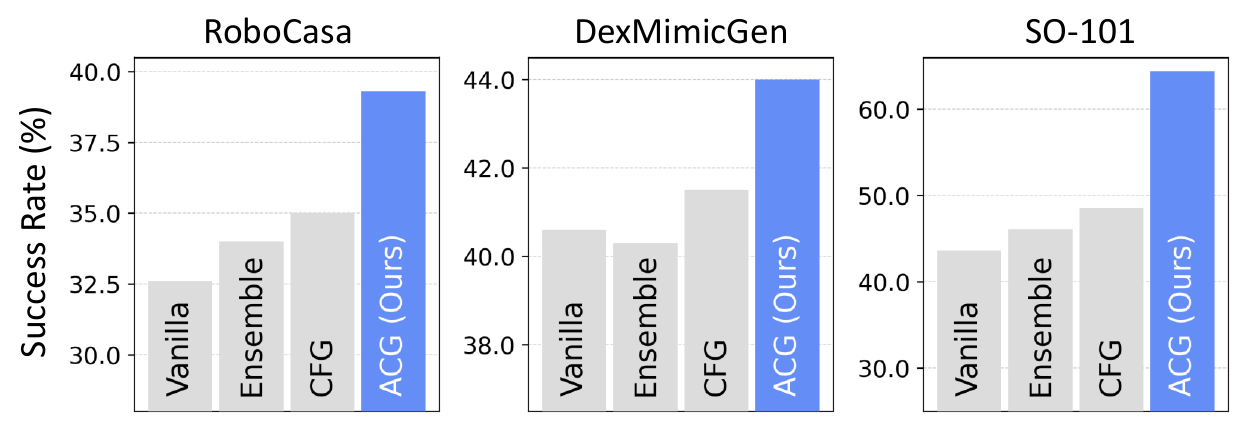}
\caption{
Performance improvements of ACG over Vanilla GR00T-N1~\cite{bjorck2025gr00t}, na\"ive multi-inference ensembling, and classifier-free guidance~\cite{ho2022classifier} on simulated manipulation benchmarks (RoboCasa~\cite{robocasa}, DexMG~\cite{jiang2025dexmimicgen}) and real-world pick-and-place tasks (SO-101~\cite{so101}).
}
\label{fig:1_teaser_performance}
\vspace{0.0cm}
\end{figure}

To address this challenge, we leverage guidance strategies from the flow matching literature, which improve prediction quality without additional training.
A representative example is Classifier-Free Guidance (CFG)~\cite{ho2022classifier}, which has been widely used in image and video generation.
CFG strengthens conditioning signals by guiding the samples away from the unconditional generation direction.
Recent studies extend the CFG by perturbing the predicted denoising vector and reversing it as guidance, thereby steering the sampling process away from undesirable characteristics and improving the fidelity of generated outputs~\cite{karras2024guiding, hyung2025spatiotemporal, ahn2024self, hong2023improving, hong2024smoothed}.

Building on these insights,  we propose \textbf{A}ction \textbf{C}oherence \textbf{G}uidance (\textbf{\method}), a simple yet effective test-time guidance strategy that enhances action coherence in flow-based policy. 
As illustrated in \Cref{fig:1_conceptual_figure}, we begin by constructing a denoising vector that drives the policy toward incoherent actions.
Specifically, \method disrupts temporal communication by replacing the attention map in the self-attention layers with an identity attention map. 
This disruption breaks coordination among tokens, yielding temporally incoherent action sequences.
\method then guides the sampling in the opposite direction of this incoherent denoising vector, encouraging temporal consistency.
As a result, \method generates action sequences with significantly enhanced coherence.

We evaluate \method on standard multi-task manipulation benchmarks, RoboCasa~\cite{robocasa} and DexMimicGen~\cite{jiang2025dexmimicgen}, as well as on real-world pick-and-place tasks using the SO-101 robot~\cite{so101}.
As shown in \Cref{fig:1_teaser_performance}, across all settings, \method delivers substantial performance gains, particularly on fine manipulation tasks such as pressing buttons (+23.1\%), insertion (+11.8\%), and real-world pick-and-place (+28.8\%).
These results demonstrate the effectiveness of \method in generating coherent action sequences with flow-based VLA models, without requiring additional training.

\section{Related Work}
\vspace{-0.03cm}

\subsection{VLA Models with Flow Matching Policy}
\vspace{-0.05cm}

Manipulation has been approached through imitation learning from human demonstrations~\cite{chi2023diffusion,pomerleau1988alvinn,mandlekar2021matters,shafiullah2022behavior,florence2022implicit,zhang2018deep,florence2019self,act}, and the research has shifted toward building foundation models, known as Vision-Language-Action (VLA) models, to enable generalization across various tasks under a single framework.
While the early VLA architectures relied on autoregressive LLM architecture~\cite{kim2025openvla,ghosh2024octo,brohan2022rt,zitkovich2023rt}, it has been shifted to employ diffusion and flow matching policy action heads~\cite{bjorck2025gr00t, black2024pi_0, shukor2025smolvla, liu2025rdt}.
VLA models take multimodal inputs, including visual observations, language instructions, states, and leverage them to predict corresponding actions.
Their architecture typically consists of a vision-language backbone coupled with an action head, where the latter can be autoregressive or generative.
In this work, we focus on improving generative flow-based VLA models (\eg, GR00T-N1~\cite{bjorck2025gr00t}, $\pi_0$~\cite{black2024pi_0}, SmolVLA~\cite{shukor2025smolvla}), which offer improved stability and expressiveness for continuous control in manipulation tasks.

\subsection{Guidance for Flow Matching Policy}
\vspace{-0.05cm}

Guidance is a widely used technique in diffusion and flow-based generative models to improve sample quality~\cite{dhariwal2021diffusion, ho2022classifier}.
A prominent example is Classifier-Free Guidance (CFG)~\cite{ho2022classifier}, which enhances text adherence by steering generation away from the unconditional vector field. 
Inspired by CFG, recent studies have explored guidance strategies in robot control, showing performance improvements in goal-conditioned imitation learning by generating negative guidance through removing the goal condition ~\cite{reuss2023goal, cfgrl, pearce2023imitating}.
However, in VLA models, applying CFG by replacing the language condition often shows unstable behaviors, since action distribution can vary significantly with even subtle differences in language instructions~\cite{liu2025rdt, liu2025hybridvla}.

Guiding the model with language condition also poses challenges in visual generation.
In particular, CFG often overly enforces the text condition, leading to the generation of unrealistic samples and reduced sample diversity.
To address this, recent work has explored perturbation guidance, which steers pretrained diffusion models toward higher quality by leveraging an intentionally degraded version of the model rather than the unconditional model~\cite{karras2024guiding}.
Such degradation can be conducted by dropping out units~\cite{karras2024guiding,hyung2025spatiotemporal}, or perturbing the attention maps~\cite{ahn2024self, hong2023improving, hong2024smoothed}.

\subsection{Action Coherent Policy}
\vspace{-0.05cm}

A straightforward remedy is to enforce temporal consistency in the generated action sequence.
For instance, smoothing the action sequence with a Gaussian kernel can generate smoother actions but at the cost of distorting the pretrained action distribution.

\paragraph{Action Chunking for Coherent Generation}
Prior work improves coherence through action chunking~\cite{act, chi2023diffusion}, which generates multiple actions (typically $k$ steps at a time) simultaneously.
By shortening the effective task horizon by a factor of $k$, action chunking reduces compounding error and promotes smoother trajectories~\cite{shi2023waypoint, belkhale2023hydra, liu2025bidirectional, malhotra2025self, black2025real, son2025lipo}. ACT~\cite{act} further proposed temporal ensembling, which can be effectively used with autoregressive architectures, but applying it to flow matching policies incurs substantial inference overhead due to their non-autoregressive nature.

While chunking enhances coherence across timesteps, it does not eliminate the incoherence that remains within each chunk.
Shaky or unstable motions can still occur within each action chunk, often leading to critical mistakes at key moments, such as grasping or picking up the objects.
To address this limitation, we aim to improve action chunking by reducing incoherence \textit{within each generated action chunk}.

% fig:2_method_overview
\begin{figure*}[t]
\centering
\includegraphics[width=0.95\linewidth]{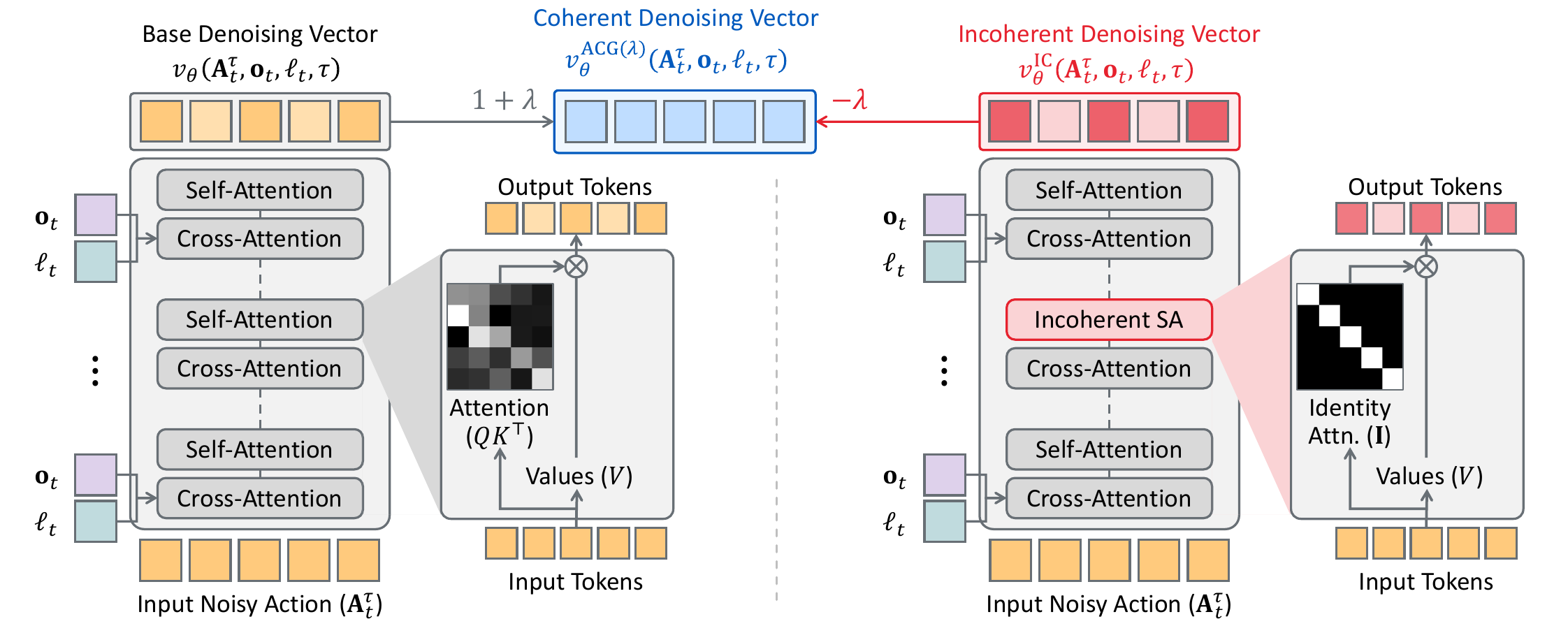}
\caption{Illustration of \method. Based on the original inference procedure (left), we modify the self-attention layer by replacing the attention map with the identity map to generate incoherent action sequence (right). Finally, we guide the denoising vector with the opposite direction of the incoherent denoising vector.}
\label{fig:2_method_overview}
\vspace{-0.4cm}
\end{figure*}

\paragraph{Guidance for Action Coherence}

More recently, researchers have explored guidance strategies to improve action coherence, motivated by the inference overhead of temporal ensembling~\cite{act} in flow-matching policies.  
These approaches guide the generation to be more consistent with the preceding action chunks~\cite{liu2025bidirectional, malhotra2025self, black2025real}.
Nevertheless, ensuring coherent actions within each chunk remains underexplored, despite its importance for fine-grained manipulation tasks.
To the best of our knowledge, this work is the first to (i) introduce perturbation guidance into robot control and (ii) explicitly address action coherence within each action chunk.

\section{Preliminaries}

\subsection{Flow Matching Policy}

Diffusion and flow matching policies have recently emerged as a powerful framework for imitation learning, enabling robots to capture the stochastic and multi-modal characteristics of human demonstrations~\cite{chi2023diffusion, black2024pi_0, bjorck2025gr00t, shukor2025smolvla}.
Here, we outline the training procedure and output representation of a rectified flow matching policy.
Human demonstration dataset can be denoted as $\{D_i\}_{i=1}^N$, where $N$ is the number of demonstrations and each demonstration $D$ is given by
\begin{equation}
    D = \left\{(\ell_1, \rvo_1, \rva_1), (\ell_2, \rvo_2, \rva_2), \dots, (\ell_T, \rvo_T, \rva_T)\right\},
\end{equation}
where $\ell_t$ denotes a language instruction and $\rvo_t$ represents the observation at time $t$.
The observation consists of multiple RGB images and the robot’s joint state.
The goal is then to model the action distribution $p(\rmA_t \mid \rvo_t, \ell_t)$, where $\rmA_t = [\rva_t, \rva_{t+1}, \dots, \rva_{t+k-1}]$ denotes an \textit{action chunk}.

During training, the noisy input action chunk $\rmA_t^\tau$ is constructed from a clean action chunk $\rmA_t$, a flow matching timestep $\tau \in [0, 1]$, and sampled noise $\epsilon \sim \gN(\mathbf{0}, \rmI)$, such that $\rmA_t^\tau = \tau \rmA_t + (1-\tau)\epsilon$.  
The conditional flow matching policy $v_\theta(\rmA_t^\tau, \rvo_t, \ell_t, \tau)$ is trained to match the conditional denoising vector field $u(\rmA_t^\tau \mid \rmA_t)$,
\begin{equation}
    u(\rmA_t^\tau \mid \rmA_t) = \frac{d\rmA_t^\tau}{d\tau} = \rmA_t - \epsilon,
\end{equation}
where subscripts denote robot timesteps and superscripts denote flow matching timesteps.  
The training objective is given by the following loss function~\cite{lipman2023flow, liu2022flow}:
\begin{equation}
    \gL (\theta) = \E_{(\ell_t, \rvo_t, \rmA_t), \tau} \big[\norm{v_\theta(\rmA_t^\tau, \rvo_t, \ell_t, \tau) - (\rmA_t - \epsilon)}^2\big].
\end{equation}

During inference, the flow matching policy generates an action chunk by integrating the learned vector field, starting from random noise $\rmA_t^0 \sim \gN(\bm{0}, \rmI)$.  
The action chunk is generated via forward Euler integration as follows:
\begin{equation}
    \rmA_t^{\tau+\delta} = \rmA_t^{\tau} + \delta \, v_\theta(\rmA_t^\tau, \rvo_t, \ell_t, \tau),
\end{equation}
where $\delta$ is the integration step size, which is set to $1/16$ in our experiments (\ie, 16 denoising timesteps)~\cite{bjorck2025gr00t}.

\subsection{Classifier-Free Guidance for Flow Matching Policy}

Classifier-Free Guidance (CFG)~\cite{ho2022classifier} has been widely used to improve conditioning in visual generation, and it has also been explored for policy networks~\cite{cfgrl, reuss2023goal, pearce2023imitating}.  
CFG is applied at inference time and defines a guided distribution as
\begin{equation}
\hspace{-0.1cm} \pi^\text{CFG($\lambda$)}(\rmA_t|\rvo_t, \ell_t) \propto \pi_\theta(\rmA_t|\rvo_t, \ell_t) \left( \frac{\pi_\theta(\rmA_t|\rvo_t, \ell_t)}{\pi_\theta(\rmA_t|\rvo_t, \emptyset)} \right)^\lambda,
\end{equation}
where $\emptyset$ denotes a null text, $\pi_\theta(\rmA_t|\rvo_t, \emptyset)$ indicates the unconditional distribution, and $\lambda$ is the guidance scale. 
Here, following the Goal-Conditioned Behavioral Cloning (GCBC) setting commonly adopted in recent flow matching policies, only the language instruction $\ell_t$ is regarded as the CFG condition, while the observation $\rvo_t$ is not~\cite{cfgrl, reuss2023goal}.
This formulation shows that CFG amplifies the contribution of the conditional distribution while pushing the sampling process away from the unconditional distribution.

To achieve this, flow-based generative models utilize the following guided vector fields~\cite{song2021score,zheng2023guided}:
\begin{align}
\label{eq:cfg}
    v_\theta^{\text{CFG($\lambda$)}}(\rmA_t^\tau, \rvo_t, \ell_t, \tau) &= (1 + \lambda)\,v_\theta(\rmA_t^\tau, \rvo_t, \ell_t, \tau) \\ \nonumber
    &\quad - \lambda\,v_\theta(\rmA_t^\tau, \rvo_t, \emptyset, \tau),
\end{align}
where $v_\theta(\rmA_t^\tau, \rvo_t, \ell_t, \tau)$ and $v_\theta(\rmA_t^\tau, \rvo_t, \emptyset, \tau)$ are the conditional and unconditional vector fields, respectively.
In summary, CFG strengthens conditional generation by pushing the denoising direction away from the unconditional vector field and toward the conditional vector field.

% algo:method
\begin{algorithm}[t]
\SetKwInOut{KwIn}{Input}
\SetKwInOut{KwOut}{Output}
\caption{Action Coherence Guidance (ACG)}
\label{algo:method}

\KwIn{observation $\rvo_t$, instruction $\ell_t$, guidance scale $\lambda$, denoising step size $\delta$}
\KwOut{generated action chunk $\rmA_t^1$}
\vspace{0.1cm}

$\rmA_t^0 \sim \mathcal{N}(\mathbf{0}, \rmI)$ \Comment{Initialize with Gaussian noise}

\For{\rm{$\tau = 0$ to 1 with step size $\delta$}}{
$v^\text{original} \gets v_\theta(\rmA_t^\tau, \rvo_t, \ell_t, \tau)$ \Comment{Infer original vector}

$v^\text{IC} \gets v_\theta^\text{IC}(\rmA_t^\tau, \rvo_t, \ell_t, \tau)$ \Comment{Infer incoherent vector}

$v_\theta^{\text{\method}} \gets (1 + \lambda)\,v^\text{original} - \lambda\,v^\text{IC}$ \Comment{Guide toward opposite}

$\rmA_t^{\tau+\delta} \gets \rmA_t^\tau + \delta \, v_\theta^{\text{\method}}$ \Comment{Denoise action chunk}
}

\Return{$\rmA_t^{1}$}
\end{algorithm}

% fig:3_exps_setup
\begin{figure*}[t]
\centering
\includegraphics[width=\linewidth]{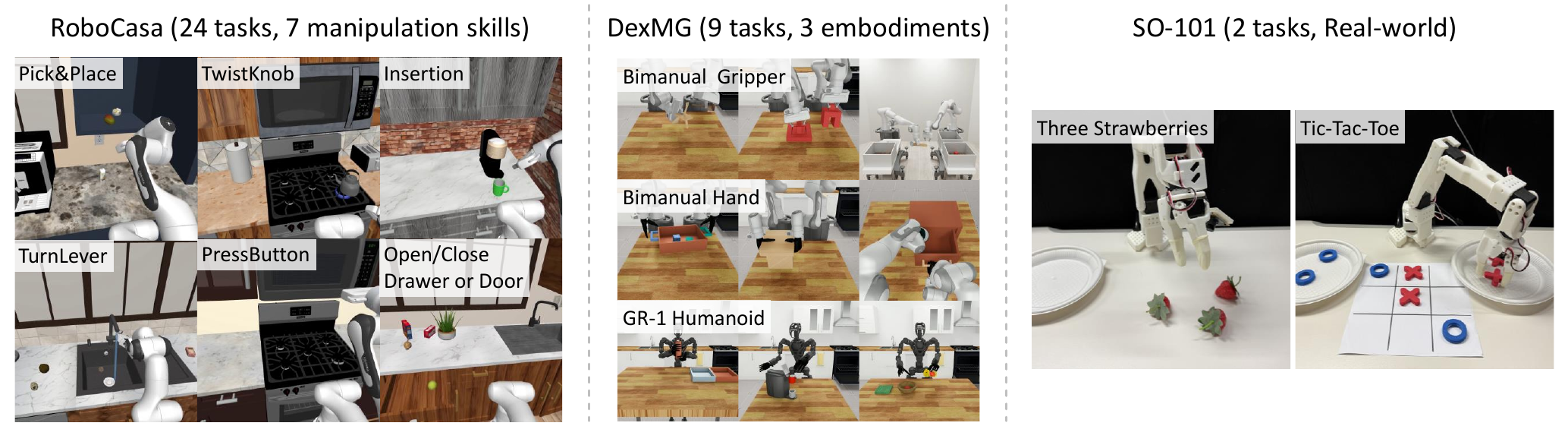}
\caption{
Visual demonstration of the simulation and real-world benchmarks used in our experiments.
RoboCasa~\cite{robocasa} contains 24 atomic manipulation tasks in kitchen environments, grouped into seven skill domains.
DexMG~\cite{jiang2025dexmimicgen} provides three bimanual embodiments for dexterous manipulation tasks.
SO-101~\cite{so101} features real-world pick-and-place tasks, including strawberries and tic-tac-toe.
}
\label{fig:3_exps_setup}
\end{figure*}

\section{Method}

\subsection{Guiding Policy with Incoherent Action Generation}

Noisy patterns such as jerks, hesitations, and overshooting in human demonstrations introduce incoherence in flow-based imitation learning, which is a critical issue for robotic manipulation tasks that demand precise object interactions. 
To address this, we propose a training-free \textit{coherent action generation} method that steers the policy away from an intentionally constructed \textit{incoherent vector field}. 
In particular, we contrast the original policy with its incoherent variant, guiding the model away from the unstable and temporally incoherent behaviors and toward stable and coherent actions, as illustrated in \Cref{fig:1_conceptual_figure}.

To achieve this, we define the guided distribution of coherent action generation as 
\begin{equation}
\pi^\text{ACG($\lambda$)}(\rmA_t|\rvo_t, \ell_t) 
\propto \pi_\theta(\rmA_t|\rvo_t, \ell_t)
\left( \frac{\pi_\theta(\rmA_t|\rvo_t, \ell_t)}{\pi_\theta^\text{IC}(\rmA_t|\rvo_t, \ell_t)} \right)^{\lambda},
\end{equation}
where $\pi_\theta^\text{IC}(\rmA_t|\rvo_t, \ell_t)$ denotes the distribution induced by the incoherent denoising vector. 
Correspondingly, the vector field for coherent action generation is formulated as:
\begin{align}
\label{eq:acg}
    v_\theta^{\text{ACG($\lambda$)}}(\rmA_t^\tau, \rvo_t, \ell_t, \tau) 
    &= (1 + \lambda)\,v_\theta(\rmA_t^\tau, \rvo_t, \ell_t, \tau) \\ \nonumber
    &\quad\,\, - \lambda\,v_\theta^\text{IC}(\rmA_t^\tau, \rvo_t, \ell_t, \tau),
\end{align}
where $\lambda$ is the guidance scale and $v_\theta^\text{IC}(\rmA_t^\tau, \rvo_t, \ell_t, \tau)$ denotes the incoherent vector field.
The specific design of the vector field is described in the following subsection.
The action chunk is then generated through forward Euler integration, in the same way as in CFG:
\begin{equation}
    \rmA_t^{\tau+\delta} = \rmA_t^{\tau} + \delta \, v_\theta^\text{ACG($\lambda$)}(\rmA_t^\tau, \rvo_t, \ell_t, \tau).
\end{equation}
The overall procedure is summarized in \Cref{algo:method}.

\subsection{Constructing Incoherent Action Generation Vector}

To construct the incoherent action vector field, we examine the architecture of the flow matching policy.
The flow matching policy adopts a fully transformer-based architecture~\cite{peebles2023scalable}, which employs self-attention to ensure coherence across generated tokens by allowing tokens to communicate with one another.
In a flow matching policy, each token represents an action at a specific timestep along the temporal horizon.
The attention operation is expressed as:
\begin{equation}
    \mathrm{Attn}(Q, K, V) 
    = \underbrace{\mathrm{softmax}\left(\frac{Q K^\top}{\sqrt{d}}\right)}_{\text{Attention Map}}
      \;\;V,
\end{equation}
where the attention map controls temporal coherence, how strongly each action attends to others, thereby promoting temporal coherence among the value tokens $V$.

Based on the attention mechanism, we construct $v_\theta^\text{IC}$ as an incoherent action generation vector field by replacing the attention operation with \textit{incoherent self-attention}.  
Concretely, the attention map is replaced with an identity attention map, forcing each action token to attend only to itself, which can be written as:
\begin{equation}
    \mathrm{Attn}_{\text{IC}}(Q,K,V) = \underbrace{\rmI}_{\text{Identity Attention Map}} \hspace{-0.7cm} V = V,
\end{equation}
where $\rmI$ indicates the identity matrix, enforcing temporal disconnection across the value tokens.
As a result, $v_\theta^\text{IC}$ produces an action chunk with reduced temporal coherence and serves as a useful reference for the opposite direction of coherent action generation, as illustrated in \Cref{fig:2_method_overview}.

% tab:3_exps_main
\begin{table*}[t]
\centering
\vspace{0.2cm}
\caption{Quantitative Comparison across Simulation and Real-world Benchmarks.}
\vspace{-0.1cm}
\label{tab:3_exps_main}
\begin{tabular}{L{3cm}|M{2.2cm}M{2.2cm}M{2.2cm}M{2.2cm}|M{2.2cm}}

\toprule
\multirow{2}{*}[-0.1cm]{Method} & \multicolumn{2}{c}{Simulation} & \multicolumn{2}{c|}{Real-world} & \multirow{2}{*}[-0.1cm]{Average} \\
\cmidrule(lr){2-3} \cmidrule(lr){4-5}
 & RoboCasa & DexMG\protect\footnotemark$^\dag$ & Three Strawberries & Tic-Tac-Toe &  \\
\midrule
Vanilla GR00T-N1 & 32.6\% {\scriptsize (±2.07\%)} & 40.6\% {\scriptsize (±3.08\%)} & 43.6\% {\scriptsize (±5.29\%)} & 38.3\% {\scriptsize (±2.89\%)} & 38.8\% {\scriptsize (±2.34\%)} \\
\midrule \multicolumn{6}{l}{\textit{Action Smoothing Methods}} \\
Ensemble ($n=2$) & 34.0\% {\scriptsize (±0.62\%)} & 40.3\% {\scriptsize (±4.42\%)} & 56.7\% {\scriptsize (±6.67\%)} & 45.0\% {\scriptsize (±5.00\%)} & 44.0\% {\scriptsize (±2.56\%)} \\
Ensemble ($n=5$) & 33.9\% {\scriptsize (±0.71\%)} & 40.0\% {\scriptsize (±5.79\%)} & 54.4\% {\scriptsize (±1.92\%)} & 43.3\% {\scriptsize (±5.77\%)} & 42.9\% {\scriptsize (±2.93\%)} \\
Action Smoothing & 34.0\% {\scriptsize (±1.40\%)} & 41.2\% {\scriptsize (±3.04\%)} & 47.8\% {\scriptsize (±1.92\%)} & 36.7\% {\scriptsize (±2.89\%)} & 39.9\% {\scriptsize (±1.55\%)} \\
Feature Smoothing & 34.4\% {\scriptsize (±1.15\%)} & 42.4\% {\scriptsize (±4.75\%)} & 57.8\% {\scriptsize (±3.85\%)} & 45.0\% {\scriptsize (±5.00\%)} & 44.9\% {\scriptsize (±1.07\%)} \\
\midrule \multicolumn{6}{l}{\textit{Guidance-based Action Generation Methods}} \\
CFG & 35.0\% {\scriptsize (±1.35\%)} & 41.5\% {\scriptsize (±3.08\%)} & 50.0\% {\scriptsize (±3.33\%)} & 43.3\% {\scriptsize (±2.89\%)} & 42.5\% {\scriptsize (±1.34\%)} \\
WNG & 35.0\% {\scriptsize (±0.47\%)} & 42.0\% {\scriptsize (±3.54\%)} & 65.6\% {\scriptsize (±5.09\%)} & 48.3\% {\scriptsize (±5.77\%)} & 47.7\% {\scriptsize (±3.34\%)} \\
\method (Ours) & \textbf{39.3\%} {\scriptsize (±3.02\%)} & \textbf{44.0\%} {\scriptsize (±2.41\%)} & \textbf{74.4\%} {\scriptsize (±3.85\%)} & \textbf{56.7\%} {\scriptsize (±2.89\%)} & \textbf{53.6\%} {\scriptsize (±0.73\%)} \\
\bottomrule

\end{tabular}
\vspace{-0.4cm}
\end{table*}

\vspace{-0.05cm}
\section{Experiments}
\vspace{-0.05cm}

We now present experiments that evaluate the effectiveness of \method for coherent action generation and, consequently, manipulation performance in VLA models. Section~\ref{sec:3_exps_evaluation} describes the experimental setup and baselines. The subsequent sections address the following research questions:
\begin{itemize}
\item How does \method improve the performance of VLA models in manipulation tasks? (\Cref{sec:3_exps_benchmark})
\item Does \method generate coherent actions? (\Cref{sec:3_exps_analysis})
\item Which self-attention layers are most effective to perturb for guidance? (\Cref{sec:3_exps_ablation})
\item How well does \method perform across various flow-based VLA models? (\Cref{sec:3_exps_ablation})
\end{itemize}

\subsection{Experimental Setup}
\label{sec:3_exps_evaluation}

\paragraph{Benchmarks}

We evaluate our method on two simulation benchmarks and one real-world benchmark.
The simulation benchmarks are drawn from open-source suites for tabletop manipulation, as illustrated in \Cref{fig:3_exps_setup}.
RoboCasa~\cite{robocasa} spans seven manipulation skill domains with a total of 24 tasks, while DexMimicGen~\cite{jiang2025dexmimicgen} provides diverse bimanual manipulation tasks across multiple embodiments.
For real-world evaluation, we conducted two pick-and-place tasks with SO-101~\cite{so101}: Three Strawberries and Tic-Tac-Toe.
In Three Strawberries, performance is scored by the number of strawberries placed, one (33.3\%), two (66.7\%), or three (100\%).
In Tic-Tac-Toe, performance is divided equally between picking (50\%) and placing (50\%) the pieces, following previous literature~\cite{jang2025dreamgen}. 
Every experiment is conducted three times to report standard deviation, with 24 trials for simulation and 10 trials for real-world experiments.

\paragraph{Implementation Details}

We primarily use GR00T-N1~\cite{bjorck2025gr00t} for most of our experiments.
For simulation, we perform multi-task training on RoboCasa~\cite{robocasa} and cross-embodiment training on DexMimicGen~\cite{jiang2025dexmimicgen}, using 100 demonstrations per task.
For the real-world evaluation, we conduct single-task training on SO-101 for the Three Strawberries and Tic-Tac-Toe tasks with 50 and 40 human demonstrations, respectively.
To improve grasp reliability given the hardware limitations, we additionally covered the SO-101 gripper with a rubber thimble.
All post-training of GR00T-N1 is performed with a batch size of 128, 60k iterations, and a peak learning rate of 0.0001, following the original setup~\cite{bjorck2025gr00t}. 
For \method, we replace the 4th–6th self-attention layers with incoherent ones out of the eight total layers, using a guidance scale of 3.0
All source code and datasets are available \href{https://github.com/DAVIAN-Robotics/ACG}{here}.

\paragraph{Baselines}

We compare our method with several practical baselines designed to enhance action generation.
First, we evaluate the vanilla VLA model, which does not employ any inference-time algorithm \textit{(i)}.
Next, we consider simple yet representative action smoothing methods that can enforce temporal coherence \textit{(ii, iii)}.
Lastly, we include other guidance-based action generation methods \textit{(iv, v)}.
\begin{enumerate}[label=\textit{(\roman*)}]
\item \textit{Vanilla GR00T-N1~\cite{bjorck2025gr00t}}: This baseline simply samples actions from the base flow-matching policy without any additional mechanism.
\item \textit{Ensemble}: We generate multiple action trajectories from the policy with different initial noise and average them. We report results with ensemble sizes of 2 and 5.
\item \textit{Smoothing}: We apply temporal smoothing with a Gaussian filter ($\sigma = 0.1$) to either the final action predictions or the intermediate action features before the self-attention layers.
\item \textit{Classifier-Free Guidance (CFG)~\cite{ho2022classifier}}: We finetune the vanilla model with a null text (dropout rate 0.1) to obtain an unconditional model from the conditional one, and generate actions according to \Cref{eq:cfg}.
\item \textit{White Noise Guidance (WNG)}: As an alternative to construct an incoherent denoising vector field ($v_\theta^\text{IC}$), we inject white noise ($\sigma = 1.0$) into the intermediate action features before the self-attention layers, disrupting temporal consistency across timesteps.
\end{enumerate}

\begin{figure}
\centering
\vspace{0.1cm}
\includegraphics[width=\linewidth]{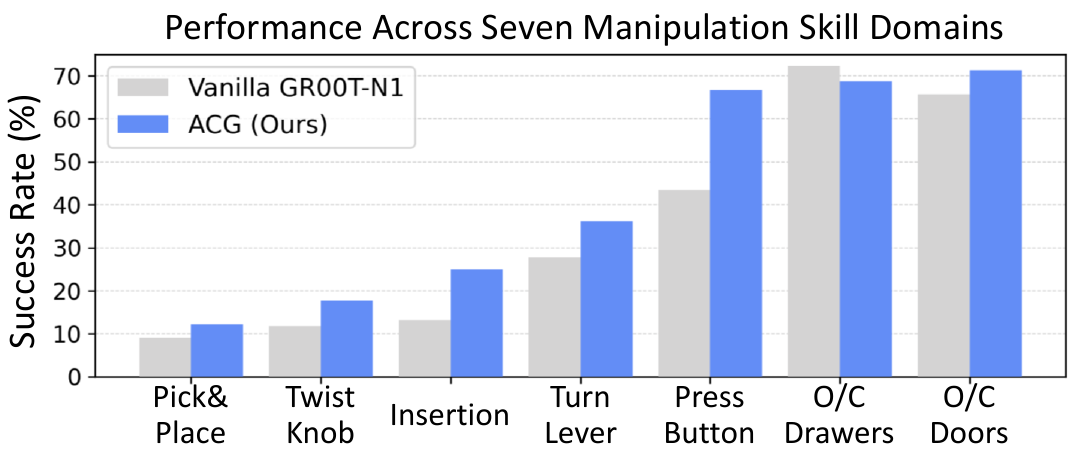}
\caption{
Performance across the seven manipulation skill domains in RoboCasa~\cite{robocasa}.  
ACG shows strong improvements on fine-grained manipulation tasks including insertion and button pressing.
}
\label{fig:3_exps_taskwise}
\end{figure}

\begin{figure*}[t]
\centering
\vspace{0.2cm}
\includegraphics[width=0.88\linewidth]{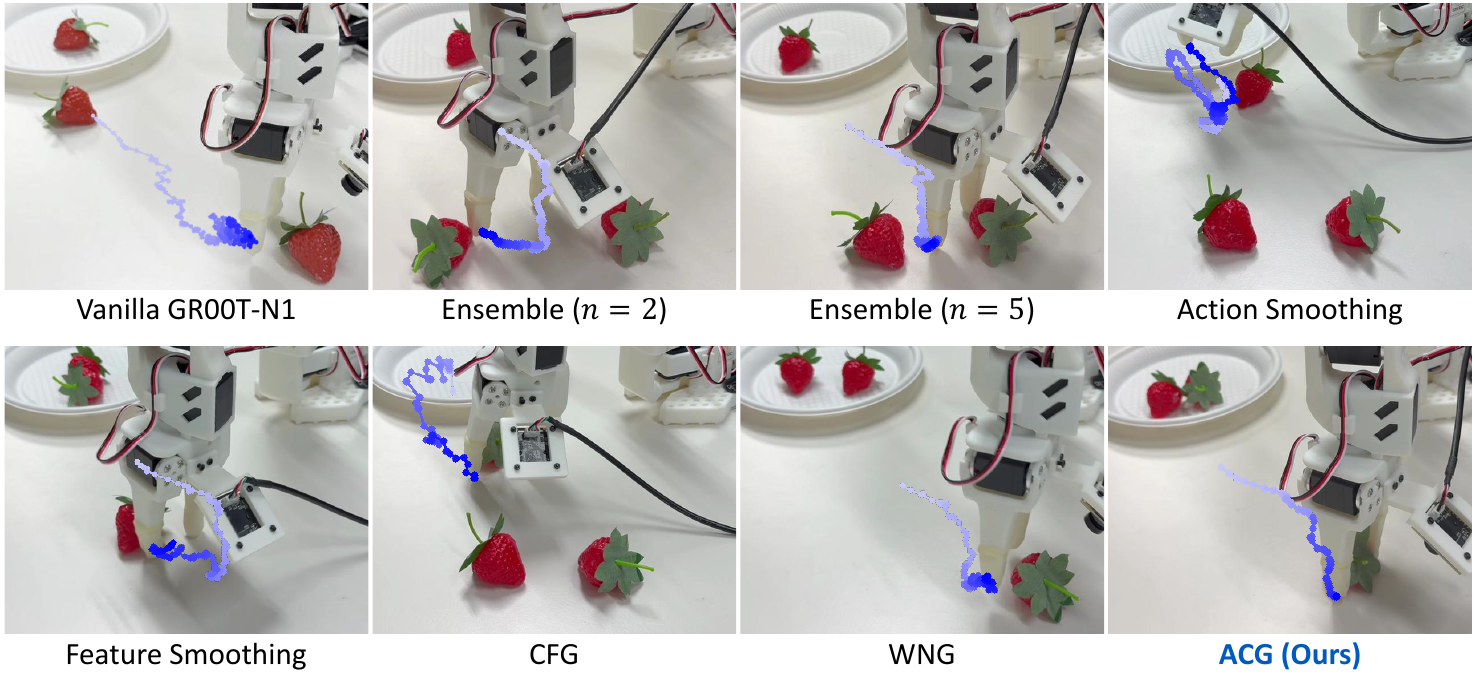}
\vspace{-0.2cm}
\caption{
Qualitative comparison on grasping strawberries.
We visualize the trajectory of the endpoint of the gripper to show stability and accuracy of the actions. The color gradient indicates temporal progression, with the trajectory becoming darker over time. While the baseline methods often generate temporally incoherent or inaccurate actions, \method generates coherent and successful trajectory.
}
\label{fig:3_exps_quali}
\vspace{-0.5cm}
\end{figure*}

\subsection{Benchmark Results}
\label{sec:3_exps_benchmark}

\footnotetext{
$^\dagger$ GR00T-N1 was finetuned on RoboCasa and DexMG with reported hyperparameters.
RoboCasa performance matched the original report, while DexMG performance was slightly lower.
}

As shown in \Cref{tab:3_exps_main}, \method consistently outperforms Vanilla GR00T-N1~\cite{bjorck2025gr00t}, a state-of-the-art VLA model, achieving gains of 6.7\% on RoboCasa, 3.4\% on DexMG, 30.8\% on the Three Strawberries task, and 9.6\% on average.
These consistent improvements across diverse benchmarks demonstrate that \method generalizes well in a plug-and-play manner without requiring additional training.

Next, we analyze other baseline approaches.
Action smoothing methods yield modest improvements over the vanilla baseline, suggesting that action smoothness is indeed important for manipulation.
However, because these methods directly smooth model features or output, they can blur fine-grained action details, leading to only marginal gains.

Guidance-based methods achieve greater improvements.
Due to the iterative nature of flow matching models, guidance can steer the model more effectively toward the desired distribution without additional tuning.
While \mbox{CFG~\cite{ho2022classifier}} strengthens text conditioning, it performs worse than other smoothing-based approaches on manipulation tasks, where coherent action generation is crucial.

WNG achieves the second-best performance after \method.
This supports our intuition that steering a flow matching model using a perturbed variant of the original denoising vector field can be effective.
However, it often exhibits a trade-off between temporal coherence and accuracy: small noise injection fails to sufficiently disrupt temporal structure, whereas large noise injection erodes pretrained knowledge.
In contrast, \method effectively disrupts temporal coherence without degrading task-relevant knowledge, achieving 4.3\% improvement over WNG on RoboCasa and 8.8\% improvement on Three Strawberries.

Task-wise performance analysis (\Cref{fig:3_exps_taskwise}) further highlights that \method is especially effective on fine-grained manipulation skills, such as button pressing (+23.1\%) and insertion (+11.8\%). 
These results indicate that action coherence is particularly crucial for fine-grained manipulation tasks. 

% tab:4_analysis_atv_jerk
\begin{table}[t]
\caption{Quantitative Action Coherence Comparison.}
\vspace{-0.1cm}
\label{tab:4_analysis_atv_jerk}
\centering
% \resizebox{\linewidth}{!}{
\begin{tabular}{l|cc}

\toprule
 &  ATV (rad/s, $\downarrow$) & JerkRMS ($\times 10^3$ rad/s$^3$, $\downarrow$) \\
\midrule
Vanilla GR00T-N1 & 1.314 {\scriptsize (±0.037)} & 1.353 {\scriptsize (±0.115)} \\
\midrule \multicolumn{3}{l}{\textit{Action Smoothing Methods}} \\
Ensemble ($n=2$) & 1.145 {\scriptsize (±0.040)} & 1.340 {\scriptsize (±0.143)} \\
Ensemble ($n=5$) & \textbf{0.984} {\scriptsize (±0.048)} & \uline{1.172} {\scriptsize (±0.128)} \\
Action Smoothing & 1.291 {\scriptsize (±0.016)} & 1.277 {\scriptsize (±0.145)} \\
Feature Smoothing & 1.287 {\scriptsize (±0.072)} & 1.233 {\scriptsize (±0.125)} \\
\midrule \multicolumn{3}{l}{\textit{Guidance-based Action Generation Methods}} \\
CFG & 1.332 {\scriptsize (±0.047)} & 1.317 {\scriptsize (±0.083)} \\
WNG & 1.274 {\scriptsize (±0.061)} & 1.265 {\scriptsize (±0.134)} \\
Incoherent ($v_\theta^\text{IC}$) & 4.509 {\scriptsize (±0.061)} & 1.993 {\scriptsize (±0.403)} \\
\method (Ours) & \uline{1.130} {\scriptsize (±0.139)} & \textbf{1.156} {\scriptsize (±0.148)} \\
\bottomrule

\end{tabular}
% }
\vspace{0.2cm}
\end{table}

\begin{figure*}
\centering
\includegraphics[width=0.98\linewidth]{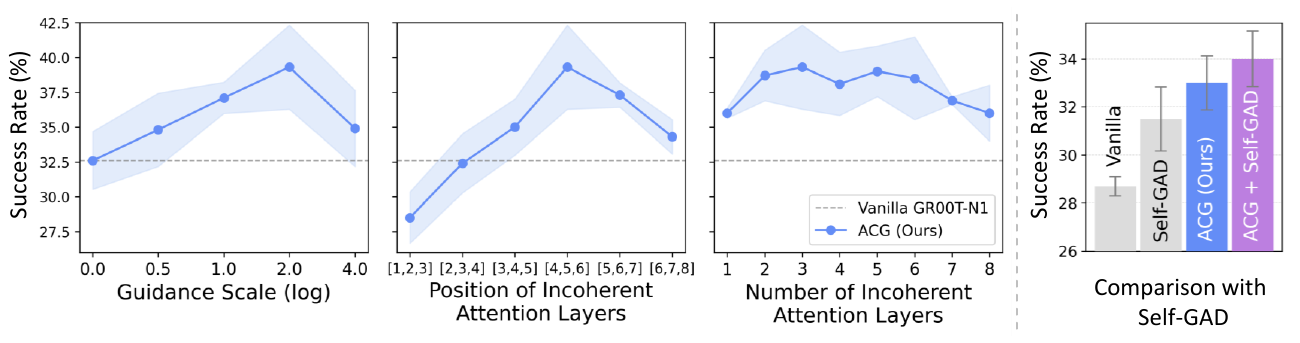}
\vspace{-0.2cm}
\captionof{figure}{
(Left) Hyperparameter analysis of \method.
Appropriate guidance scales and the number and positions of incoherent attention layers substantially improve performance.
(Right) Comparison with Self-GAD~\cite{ahn2024self} on the RoboCasa~\cite{robocasa} benchmark.
While Self-GAD enhances action coherence between action chunks, \method focuses on coherence within each action chunk.
The results suggest that intra-chunk coherence plays a more crucial role than inter-chunk coherence, and that combining both methods can yield further improvements.
}
\label{fig:4_analysis_hparams}
\vspace{-0.15cm}
\end{figure*}

\subsection{Action Coherence Analysis}
\label{sec:3_exps_analysis}

We now examine whether \method indeed improves the coherence of action sequences, quantitatively and qualitatively.

To quantitatively evaluate the action stability of the proposed method, we measure the Action Total Variation (ATV, rad/s)~\cite{rudin1992nonlinear} and JerkRMS (rad/s$^3$)~\cite{flash1985coordination}.
ATV quantifies the temporal coherence of the action sequences predicted by the flow matching policy.
Jerk, defined as the third derivative of the motor angle $\rvs$, represents the rate of change in acceleration.
JerkRMS measures the root mean square of this jerk to assess the smoothness of the resulting movements.
Formally, the metrics are defined as follows:
\begin{equation}
    \text{ATV} = \frac{1}{M(T-1)} \sum_{t=1}^{T-1}\sum_{j=1}^{M} |a_{t+1}^j - a_t^j|,
\end{equation}
\begin{equation}
    \text{JerkRMS} = \sqrt{\frac{1}{T-3} \sum_{t=1}^{T-3} \norm{\dddot{\rvs}_{t}}_2^2},
\end{equation}
where $a_{t}^j$ denotes the action of the $j$-th motor at the $t$-th timestep, and $\dddot{\rvs}_t$ denotes the jerk of the motors.
In our experiments, we report the average ATV and JerkRMS across the six motors of SO-101.
For a fair comparison, we compute the metrics only over the approach phase (\ie, the first 64 timesteps) toward a single strawberry, since later trajectories diverge depending on task success or failure.

As shown in \Cref{tab:4_analysis_atv_jerk} and \Cref{fig:3_exps_quali}, the Vanilla GR00T-N1~\cite{bjorck2025gr00t} model exhibits severe jerking and jittering, often knocking the strawberry away. 
Ensemble methods significantly improve temporal coherence; 
however, as seen in our qualitative results, they often hesitate between strawberries because the averaging process blurs distinct action paths. 
Similarly, while action and feature smoothing methods reduce action variation, they frequently produce inaccurate actions since they modify the sequence without any task-specific prior.

For guidance-based action generation methods, CFG does not improve action coherence compared to Vanilla GR00T-N1. 
These results highlight the orthogonality between improving action coherence and strengthening the goal condition in manipulation tasks. 
Based on our benchmark results (\Cref{tab:3_exps_main}), we argue that action coherence is more crucial than goal conditioning for manipulation performance. 
Lastly, while WNG can generate coherent actions similar to smoothing methods, it often sacrifices accuracy, sometimes pushing the strawberry away.

\begin{table}[]
\centering
\caption{Generalization of \method Across VLA Models}
\label{tab:3_exps_generalization}
% \resizebox{0.8\linewidth}{!}{
\begin{tabular}{l|cc}

\toprule
 & Three Strawberries & Tic-Tac-Toe \\
\midrule
$\pi_0$~\cite{black2024pi_0} & 41.1\% {\scriptsize (±5.09\%)} & 33.3\% {\scriptsize (±2.89\%)} \\
w/ \method & \textbf{53.3\%} {\scriptsize (±3.33\%)} & \textbf{46.7\%} {\scriptsize (±2.89\%)} \\
\midrule
SmolVLA~\cite{shukor2025smolvla} & 16.7\% {\scriptsize (±3.33\%)} & 23.3\% {\scriptsize (±2.89\%)} \\
w/ \method & \textbf{22.2\%} {\scriptsize (±1.92\%)} & \textbf{30.0\%} {\scriptsize (±5.00\%)}\\
\bottomrule

\end{tabular}
% }
\end{table}

As expected, the incoherent variant ($v_\theta^\text{IC}$) exhibits even worse action coherence than Vanilla GR00T-N1. 
By extrapolating this incoherent variant using \Cref{eq:acg}, \method achieves the most coherent actions, as indicated by JerkRMS, while maintaining an ATV score comparable to ensemble methods. 
Unlike ensemble methods, which often produce inaccurate action sequences, \method generates accurate and temporally consistent actions (\Cref{fig:3_exps_quali}), leading to higher success rates. 
In summary, \method produces highly coherent action sequences without sacrificing accuracy, thereby substantially improving manipulation performance.

\subsection{Ablation Study}
\label{sec:3_exps_ablation}

We conduct an ablation study to gain deeper insights into the effectiveness and robustness of \method.

\paragraph{Hyperparameter Impact}

% We analyze three hyperparameters of \method, the guidance scale (default: 3.0), the number of incoherent self-attention layers, and their positions (default: 3 layers at positions 4, 5, and 6).  
% We vary each hyperparameter individually on the RoboCasa benchmark~\cite{robocasa} while keeping the others fixed.
% As shown in \Cref{fig:4_analysis_hparams}, increasing the guidance scale improves performance, consistent with prior findings in the classifier-free guidance literature~\cite{ho2022classifier, cfgrl}.
% However, excessively large scales degrade performance, likely due to increased deviation from the base pretrained distribution.
% Using 2--6 incoherent attention layers consistently improves performance, suggesting that \method is robust to this hyperparameter.
% For their positions, while early layers occasionally reduce performance, the middle and later layers generally yield better results.

We analyze three hyperparameters of \method: the guidance scale (default: 3.0), the number of incoherent self-attention layers, and their positions (default: layers 4--6), varying each individually on RoboCasa~\cite{robocasa}.
As shown in \Cref{fig:4_analysis_hparams}, increasing the guidance scale improves performance, consistent with prior findings~\cite{ho2022classifier, cfgrl}, though excessively large scales degrade performance due to deviation from the pretrained distribution.
Using 2--6 incoherent layers consistently improves performance, suggesting robustness to this choice.
For layer positions, middle and later layers generally yield better results, while early layers occasionally hurt performance.

\paragraph{Comparison with Self-GAD}
% Self-GAD is a concurrent work that improves cross-chunk coherence through test-time guidance.  
% It is applicable only under the receding horizon setting, where the policy predicts 16 actions but executes only the first 8.
% While the receding horizon improves reactiveness, it effectively doubles the decision horizon, which amplifies compounding errors and lowers the baseline performance.
% Although their receding horizon setting lowers the baseline performance, we used the setup when comparing with Self-GAD for fair comparison.

% While both \method and Self-GAD outperform the baseline, \method achieves higher success rates.
% This indicates that, although mitigating inter-chunk errors and enhancing intra-chunk coherence both improve performance, the latter is more critical for manipulation tasks.
% Furthermore, combining \method with Self-GAD achieves the best results, showing that our method is complementary to inter-chunk approaches and can be integrated for additional gains.

Self-GAD~\cite{malhotra2025self} is a concurrent work that improves cross-chunk coherence through test-time guidance, applicable only under the receding horizon setting where the policy predicts 16 actions but executes the first 8.
While receding horizon improves reactiveness, it effectively doubles the decision horizon, amplifying compounding errors and lowering baseline performance.
Despite this, we adopted their setup for fair comparison.
Both \method and Self-GAD outperform the baseline, but \method achieves higher success rates, indicating that intra-chunk coherence is more critical than inter-chunk consistency for manipulation.
Furthermore, combining both methods achieves the best results, showing they are complementary.

\paragraph{Generalization to Flow-based Action Generation Models}

We also evaluate our approach on various VLA models.  
Specifically, we test it with standard flow-based VLA models, including $\pi_0$~\cite{black2024pi_0} and SmolVLA~\cite{shukor2025smolvla}.  
Thanks to its simplicity, the proposed method can be readily applied to any model that incorporates self-attention layers over action tokens.  
As shown in \Cref{tab:3_exps_generalization}, \method outperforms the vanilla model with a notable margin, demonstrating its generalization capability across the model variations.

\section{Conclusion}

% We introduced \method, a test-time guidance method that steers flow policies toward generating coherent action trajectories by leveraging the negative predictions of an intentionally constructed incoherent variant of the base model. \method achieves notable gains in success rates across both simulation benchmarks (RoboCasa and DexMimicGen) and real-world tasks with SO-101, and further demonstrates robustness when applied to different DiT-based flow models.

% However, these gains come with computational costs. 
% Na\"ively generating the guidance vector requires a second forward pass through the incoherent vector field. Fortunately, this cost can be reduced below $2\times$ by reusing the intermediate features. 
% As shown in \Cref{sec:3_exps_analysis}, later attention layers contribute most to incoherence, enabling us to cache the outputs of earlier layers.
% In our base setting, we shared the first half of the layers, which reduces the computational overhead to about $1.5\times$.
% Still, it remains an open question whether perturbing only a small fraction of the latter layers suffices for deeper networks.

% We hope that \method inspires the research community to view test-time guidance not only as a tool for image and video generation, but also as a principle for generating coherent and reliable actions in robotics.
% Accordingly, exploring different guidance strategies to mitigate other drawbacks of flow matching could be a promising research direction.

We introduced \method, a test-time guidance method that steers flow policies toward generating coherent action trajectories by leveraging the negative predictions of an intentionally constructed incoherent model variant. \method achieves notable gains in success rates across simulation benchmarks (RoboCasa and DexMimicGen) and real-world tasks with SO-101, and further demonstrates robustness across different DiT-based flow models.
However, these gains come with computational costs, as generating the guidance vector requires a second forward pass. By caching earlier layer outputs and perturbing only later attention layers (\Cref{sec:3_exps_analysis}), we reduce this overhead to approximately $1.5\times$, though whether perturbing only a small fraction of layers suffices for deeper networks remains an open question.
Additionally, since \method operates within individual action chunks, extending coherence guidance across chunk boundaries is an important direction for future work.
We hope that \method inspires the community to view test-time guidance not only as a tool for generation, but also as a principle for producing coherent and reliable actions in robotics.

%                                   % the textheight of the last page by a suitable amount.
%                                   % This command does not take effect until the next page
%                                   % so it should come on the page before the last. Make
%                                   % sure that you do not shorten the textheight too much.



\section*{ACKNOWLEDGMENT}

% KETI, B200, others
% Hardware technical support by 이병근, 심재현.

This work was supported by Institute for Information \& communications Technology Planning \& Evaluation(IITP) grant funded by the Korea government(MSIT) (RS-2019-II190075, Artificial Intelligence Graduate School Program(KAIST) and RS-2025-02653113, High-Performance Research AI Computing Infrastructure Support at the 2 PFLOPS Scale), the National Research Foundation of Korea(NRF) grant funded by the Korea government(MSIT) (No. RS-2025-00555621), and the ``Advanced GPU Utilization Support Program'' funded by the Government of the Republic of Korea (Ministry of Science and ICT).

\bibliographystyle{IEEEtran}
\bibliography{IEEEabrv,reference}

\end{document}